\documentclass[10pt, conference, compsocconf]{IEEEtran}

\ifCLASSINFOpdf
\else
\fi
\usepackage[bookmarks=false]{hyperref}
\usepackage{amsmath}
\usepackage{amssymb}
\usepackage{times}
\usepackage{multicol}
\usepackage{multirow}
\usepackage{bbding}
\usepackage{booktabs}
\usepackage{algorithm}
\usepackage{algorithmic}
\usepackage{graphicx}
\usepackage{subfigure}
\usepackage{color}
\usepackage{xcolor}
\usepackage{url}
\usepackage{array}
\usepackage{bm}
\usepackage{enumitem}
\usepackage[misc]{ifsym}
\begin{document}
	\title{Transfer Learning with Dynamic Adversarial Adaptation Network}

\author{\IEEEauthorblockN{
		Chaohui Yu\IEEEauthorrefmark{1}\IEEEauthorrefmark{2},
		Jindong Wang\IEEEauthorrefmark{3},
		Yiqiang Chen\IEEEauthorrefmark{1}\IEEEauthorrefmark{2}$^($\Envelope$^)$,
		Meiyu Huang\IEEEauthorrefmark{4}
	}
	\IEEEauthorblockA{\IEEEauthorrefmark{1}Beijing Key Lab. of Mobile Computing and Pervasive Device, Inst. of Comp. Tech., Chinese Academy of Sciences, Beijing, China}
	\IEEEauthorblockA{\IEEEauthorrefmark{2}University of Chinese Academy of Sciences, Beijing, China}
	\IEEEauthorblockA{\IEEEauthorrefmark{3}Microsoft Research Asia, Beijing, China}
	\IEEEauthorblockA{\IEEEauthorrefmark{4}Qian Xuesen Lab. of Space Technology, China Academy of Space Technology, Beijing, China}
	Email:\{yuchaohui17s,yqchen\}@ict.ac.cn, jindong.wang@microsoft.com}
	
\maketitle

\begin{abstract}
\label{abstract}
The recent advances in deep transfer learning reveal that adversarial learning can be embedded into deep networks to learn more transferable features to reduce the distribution discrepancy between two domains. Existing adversarial domain adaptation methods either learn a single domain discriminator to align the global source and target distributions, or pay attention to align subdomains based on multiple discriminators. However, in real applications, the marginal (global) and conditional (local) distributions between domains are often contributing differently to the adaptation. There is currently no method to dynamically and quantitatively evaluate the relative importance of these two distributions for adversarial learning. In this paper, we propose a novel Dynamic Adversarial Adaptation Network~(DAAN) to dynamically learn domain-invariant representations while quantitatively evaluate the relative importance of global and local domain distributions. To the best of our knowledge, DAAN is the first attempt to perform dynamic adversarial distribution adaptation for deep adversarial learning. DAAN is extremely easy to implement and train in real applications. We theoretically analyze the effectiveness of DAAN, and it can also be explained in an attention strategy. Extensive experiments demonstrate that DAAN achieves better classification accuracy compared to state-of-the-art deep and adversarial methods. Results also imply the necessity and effectiveness of the dynamic distribution adaptation in adversarial transfer learning.
\end{abstract}


\IEEEpeerreviewmaketitle

\section{Introduction}
\label{sec-intro}
Deep neural networks have significantly improved the performance of diverse data mining and computer vision applications~\cite{he2016deep,krizhevsky2012imagenet}. In order to avoid 
overfitting and achieve better performance, a large amount of labeled data is needed to train a deep learning model. 
Unfortunately, it is often expensive and time-consuming to acquire sufficient labeled data. Thus, a natural idea is to leverage the abundant labeled samples in an existing domain~(i.e. \textit{source} domain) to facilitate learning in the domain to be learned (i.e. \textit{target} domain). 

A promising approach to solve such cross-domain learning problems is called \textit{transfer learning}, or \textit{domain adaptation}~\cite{pan2010survey}. The key to successful adaptation is to learn a discriminative model to reduce the distribution discrepancy between the two domains. 
Traditional methods perform adaptation by either reweighting samples from the source domain~\cite{huang2007correcting,chen2019cross,wang2019easy}, or seeking an explicit feature transformation that transforms the source and target samples into the same feature space~\cite{wang2018visual,wang2017balanced,gong2012geodesic,wang2018stratified,pan2011domain}. 
Recent studies have indicated that deep networks can learn more transferable features for domain adaptation~\cite{donahue2014decaf,yosinski2014transferable}. The latest advances have been achieved by embedding domain adaptation modules in the pipeline of deep feature learning to extract domain-invariant representations~\cite{zhang2018collaborative,zhu2019multi,yu2019accelerating,sun2016deep,ganin2014unsupervised,wang2018deep}. 

\begin{figure}[t!]
    \centering
    \includegraphics[scale=.28]{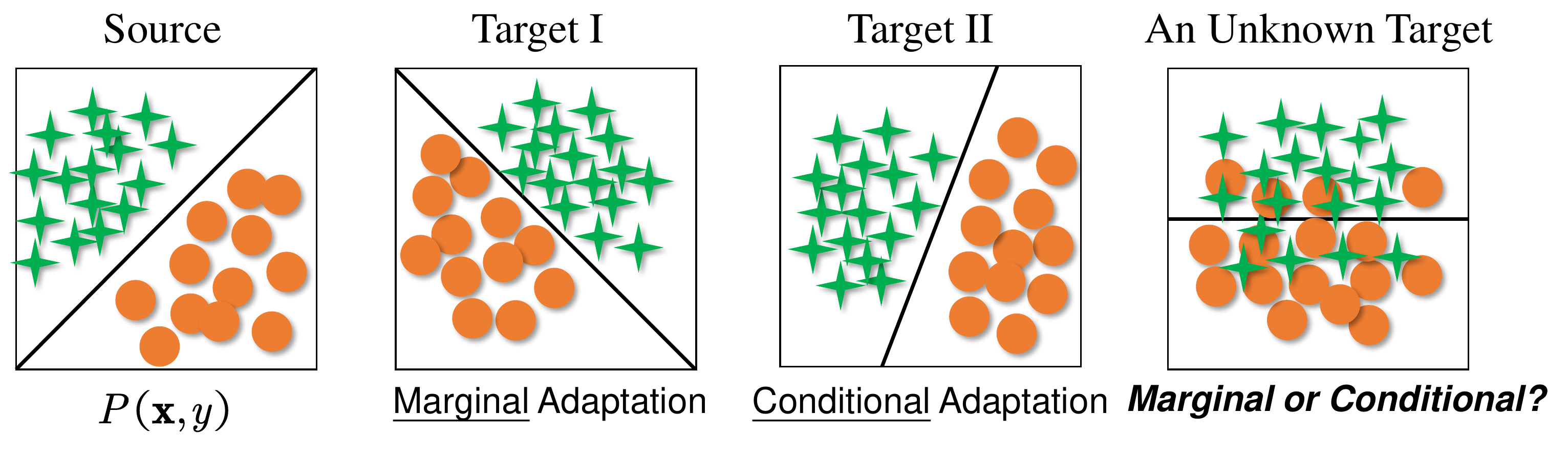}
    \caption{The different effects of marginal and conditional distributions in transfer learning applications}
    \label{fig-sub-datamc}
    \vspace{-.1in}
\end{figure}

\begin{table}[t!]
\caption{\upshape Comparison between the recent methods on Office-Home~\cite{venkateswara2017deep} dataset ($C$ denotes number of classes)}
\label{tb-com}
\resizebox{.48\textwidth}{!}{
\begin{tabular}{ccccc}
\toprule
Method & DANN~\cite{ganin2014unsupervised} & JAN~\cite{long2016deep} & MEDA~\cite{wang2018visual} & DAAN \\ \hline
Dynamic adaptation & No & No & Yes & \textbf{Yes} \\ 
Hyperparameter & $\lambda$ & $\lambda$ & $\lambda,p,\eta,\rho$ & $\lambda$ \\ 
Extra classifier & No & No & $C+1$ & \textbf{No} \\ 
Accuracy (\%) & 57.6 & 58.3 & 60.2 & \textbf{61.8} \\ \bottomrule
\end{tabular}}
\vspace{-.2in}
\end{table}

Recently, adversarial learning~\cite{goodfellow2014generative} has been successfully embedded into deep networks to reduce distribution discrepancy between the source and target domains. Prior advanced adversarial adaptation methods~\cite{ganin2014unsupervised,tzeng2017adversarial,motiian2017few,pei2018multi} have shown promising results in several domain adaptation tasks. Most of them either learn a single domain discriminator to align the global source and target distributions, or pay attention to align subdomains based on multiple discriminators. For instance, Domain-adversarial Neural Network~(DANN)~\cite{ganin2014unsupervised,ganin2016domain} focuses on the global adversarial learning, while Multi-adversarial Domain Adaptation~(MADA)~\cite{pei2018multi} pays attention to the subdomain adaptation by training several domain classifiers.
However, in real applications, the marginal (global) and conditional (local) distributions between domains are often contributing differently to the adaptation. For example, when two domains are very dissimilar~(source $\rightarrow$ target I in Fig.~\ref{fig-sub-datamc}), the global distribution is more important. When the global distributions are close (source $\rightarrow$ target II in Fig.~\ref{fig-sub-datamc}), the local distribution should be given more attention. Two more recent work called Balanced Distribution Adaptation~(BDA)~\cite{wang2017balanced} and Manifold Embedded Distribution Alignment~(MEDA)~\cite{wang2018visual} proposed to adaptively align these two distributions, while it is based on kernel method with high computational cost. In addition, MEDA is incapable of handling large-scale data. To date, there is no effort that could dynamically evaluate the relative importance of the marginal and conditional distributions for adversarial domain adaptation. 

In this paper, we propose a novel \textbf{Dynamic Adversarial Adaptation Network~(DAAN)} for unsupervised domain adaptation. DAAN is able to learn domain-invariant features through end-to-end adversarial training. The key component in DAAN is the \textit{Dynamic Adversarial Factor}, which is capable of easily, dynamically, and quantitatively evaluating the relative importance of the marginal and conditional distributions. The adaptation can be achieved by Stochastic Gradient Descent~(SGD) with the gradients computed by backpropagation in linear-time. To the best of our knowledge, DAAN is the \textit{first} adversarial domain adaptation method that is able to dynamically learn the relationship between the marginal and conditional distributions. Extensive experiments demonstrate that DAAN outperforms state-of-the-art methods on standard domain adaptation benchmarks. More importantly, it is shown that there does exist the relative importance of two distributions, of which DAAN could make accurate evaluation.

The contributions of this paper are four-fold:

(1)~We propose a novel dynamic adversarial adaptation network to learn domain-invariant features. DAAN is accurate and robust, and can be easily implemented by most deep learning libraries. 

(2)~We propose the dynamic adversarial factor to easily, dynamically, and quantitatively evaluate the relative importance of the marginal and conditional distributions in adversarial transfer learning. 

(3) We theoretically analyze the effectiveness of DAAN, and it can also be explained in an attention stragegy.

(4)~Extensive experiments on public datasets demonstrate the significant superiority of our DAAN in both classification accuracy and the estimation of the dynamic adversarial factor.


\section{Related Work}
\label{sec-related}

\subsection{Unsupervised Domain Adaptation}
Unsupervised domain adaptation is a specific area of transfer learning~\cite{pan2010survey}, which is to learn a discriminative model in the presence of the domain shift between domains. There are mainly two categories: traditional (shallow) learning and deep learning. 

Traditional (shallow) learning methods can mainly be divided into two categories: 
(1)~Subspace learning. Subspace Alignment (SA)~\cite{fernando2013unsupervised} aligns the base vectors of both domains and Subspace Distribution Alignment (SDA)~\cite{sun2015subspace} extends SA by adding the subspace variance adaptation. CORAL~\cite{sun2016return} aligns subspaces in second-order statistics. 
(2)~Distribution alignment. Pan \textit{et al.} proposed the Transfer Component Analysis~(TCA)~\cite{pan2011domain} to align the marginal distributions between domains. Based on TCA, Joint Distribution Adaptation~(JDA)~\cite{long2013transfer} is proposed to match both marginal and conditional distributions. But these works treat the two distributions equally and fail to leverage the different importance of distributions. Recently, Wang \textit{et al.} proposed Balanaced Distribution Adaptation~(BDA)~\cite{wang2017balanced} and Manifold Embedded Distribution Alignment~(MEDA)~\cite{wang2018visual} approaches to dynamically evaluate the different effect of marginal and conditional distributions and achieved the state-of-the-art results on domain adaptation. However, MEDA is based on kernel method and requires to train several linear classifiers in each iteration. DAAN is significantly different from MEDA in two folds as shown in Table~\ref{tb-com}. Firstly, MEDA uses shallow features to learn the adaptive factor, while DAAN uses deep adversarial representations for end-to-end learning. Secondly, MEDA uses extra linear classifiers to learn the adaptive factor, while DAAN directly uses the adversarial features, which is more efficient.

In recent years, deep networks can learn more transferable features for domain adaptation~\cite{donahue2014decaf,yosinski2014transferable}, by disentangling explanatory factors of variations behind domains compared to traditional methods. 
Most work on deep domain adaptation is based on discrepancy measurement. For instance, Correlation Alignment (CORAL)~\cite{sun2016deep}, Kullback-Leibler (KL) divergence~\cite{zhuang2015supervised}, Maximum Mean Discrepancy (MMD)~\cite{long2015learning,long2016deep,tzeng2014deep,yan2017mind,zhu2019multi}, and Central Moment Discrepancy (CMD)~\cite{zellinger2017central} are used to reduce the distribution divergence between domains. However, there is no effective deep learning method that can dynamically align the marginal and conditional distributions.

\subsection{Domain-adversarial Learning}
As a special case of deep domain adaptation, domain-adversarial learning has been popular in recent years. In this case, a domain discriminator that classifies whether a data point is drawn from the source or target domain is used to encourage domain confusion through an adversarial objective to minimize the distance between the source and target distributions~\cite{ganin2014unsupervised}. 
Adversarial learning has been explored in Generative Adversarial Networks~(GANs)~\cite{goodfellow2014generative}. And Generative Multi-Adversarial Network (GMAN)~\cite{durugkar2016generative} extends GANs to multiple discriminators including formidable adversary and forgiving teacher, which significantly eases model training.

Recently, we have witnessed considerable research~\cite{ganin2014unsupervised,pei2018multi,kumar2018co,xie2018learning} for adversarial domain adaptation. 
DANN~\cite{ganin2014unsupervised} aligns the whole source and target distributions with a global domain discriminator. MADA~\cite{pei2018multi} captures multi-mode structures to enable fine-grained alignment of different data distributions based on multiple domain discriminators. Co-DA~\cite{kumar2018co} constructs multiple diverse feature spaces and aligns source and target distributions in each of them individually. The proposed DAAN is also based on adversarial learning, while it significantly outperforms existing methods by dynamically evaluating the relative importance of the marginal and conditional distributions.

\section{Dynamic Adversarial Adaptation Network}
\label{sec-method}
In this section, we introduce the proposed Dynamic Adversarial Adaptation Network~(DAAN).

\subsection{Problem Definition}
\label{sec-problem}
In unsupervised domain adaptation, we are given a source domain $\mathcal{D}_{s}=\{(\mathbf{x}^{s}_{i},y^{s}_{i})\}^{n_s}_{i=1}$ of $n_{s}$ labeled examples and 
a target domain $\mathcal{D}_{t}=\{\mathbf{x}^{t}_{j}\}^{n_t}_{j=1}$ of $n_{t}$ unlabeled examples. $\mathcal{D}_{s}$ and $\mathcal{D}_{t}$ have the same label space, i.e. $\mathbf{x}_{i}, \mathbf{x}_{j} \in \mathbb{R}^d$ where $d$ is the dimensionality. The marginal distributions between two domains are different, i.e. $P_s(\mathbf{x}_s) \ne P_t(\mathbf{x}_t)$. The goal of deep UDA is to design a deep neural network that enables learning of transfer classifiers $y = f(\mathbf{x})$ that formally reduces the shifts in the distributions of two domains such that the target risk $\epsilon_{t}(f) = \mathbb{E}_{(x,y) \sim q}[f(x) \neq y]$ can be bounded by using the source domain while achieving better performance on the target domain.

\subsection{Adversarial Learning for Domain Adaptation}

Domain adversarial adaptation methods borrow the idea of GAN~\cite{goodfellow2014generative} to help learn transferable features. 
The adversarial learning procedure is a two-player game, where the first player is the domain discriminator $G_{d}$ trained to distinguish the source domain from the target domain, and the second player is the feature extractor $G_{f}$ that tries to confuse the domain discriminator by extracting domain-invariant features. The two players are trained adversarially: the parameters $\theta_f$ of feature extractor $G_f$ are learned by maximizing the loss of domain discriminator $G_d$, while the parameters $\theta_d$ of $G_d$ are trained by minimizing the loss of the domain discriminator. In addition, the loss of the label classifier $G_y$ is also minimized. The loss function can be formalized as:
\begin{equation}
    \label{eq-gan}
    \begin{aligned} L\left(\theta_{f}, \theta_{y}, \theta_{d}\right)=& \frac{1}{n_{s}} \sum_{\mathbf{x}_{i} \in \mathcal{D}_{s}} L_{y}\left(G_{y}\left(G_{f}\left(\mathbf{x}_{i}\right)\right), y_{i}\right) \\ &-\frac{\lambda}{n_s+n_t} \sum_{\mathbf{x}_{i} \in\left(\mathcal{D}_{s} \cup \mathcal{D}_{t}\right)} L_{d}\left(G_{d}\left(G_{f}\left(\mathbf{x}_{i}\right)\right), d_{i}\right), \end{aligned}
\end{equation}
where $\lambda$ is a trade-off parameter and $L_{y}$ and $L_{d}$ denote the label classifier loss and domain discriminator loss. Since there are no labels for the target domain, $d_{i}$ means the domain label of the input samples~($d_{i}$ for source domain is $0$, $d_{i}$ for target domain is $1$). After the training converges, the parameters $\hat{\theta}_{f}, \hat{\theta}_{y}, \hat{\theta}_{d}$ will deliever a saddle point of Eq.~(\ref{eq-gan}):
\begin{equation}
\label{eq-argmin}
\begin{array}{c}{\left(\hat{\theta}_{f}, \hat{\theta}_{y}\right)=\arg \min _{\theta_{f}, \theta_{y}} L\left(\theta_{f}, \theta_{y}, \theta_{d}\right)} \\ {\left(\hat{\theta}_{d}\right)=\arg \max _{\theta_{d}} L\left(\theta_{f}, \theta_{y}, \theta_{d}\right)}\end{array}
\end{equation}

\begin{figure*}[t!]
	\centering\includegraphics[scale=0.35]{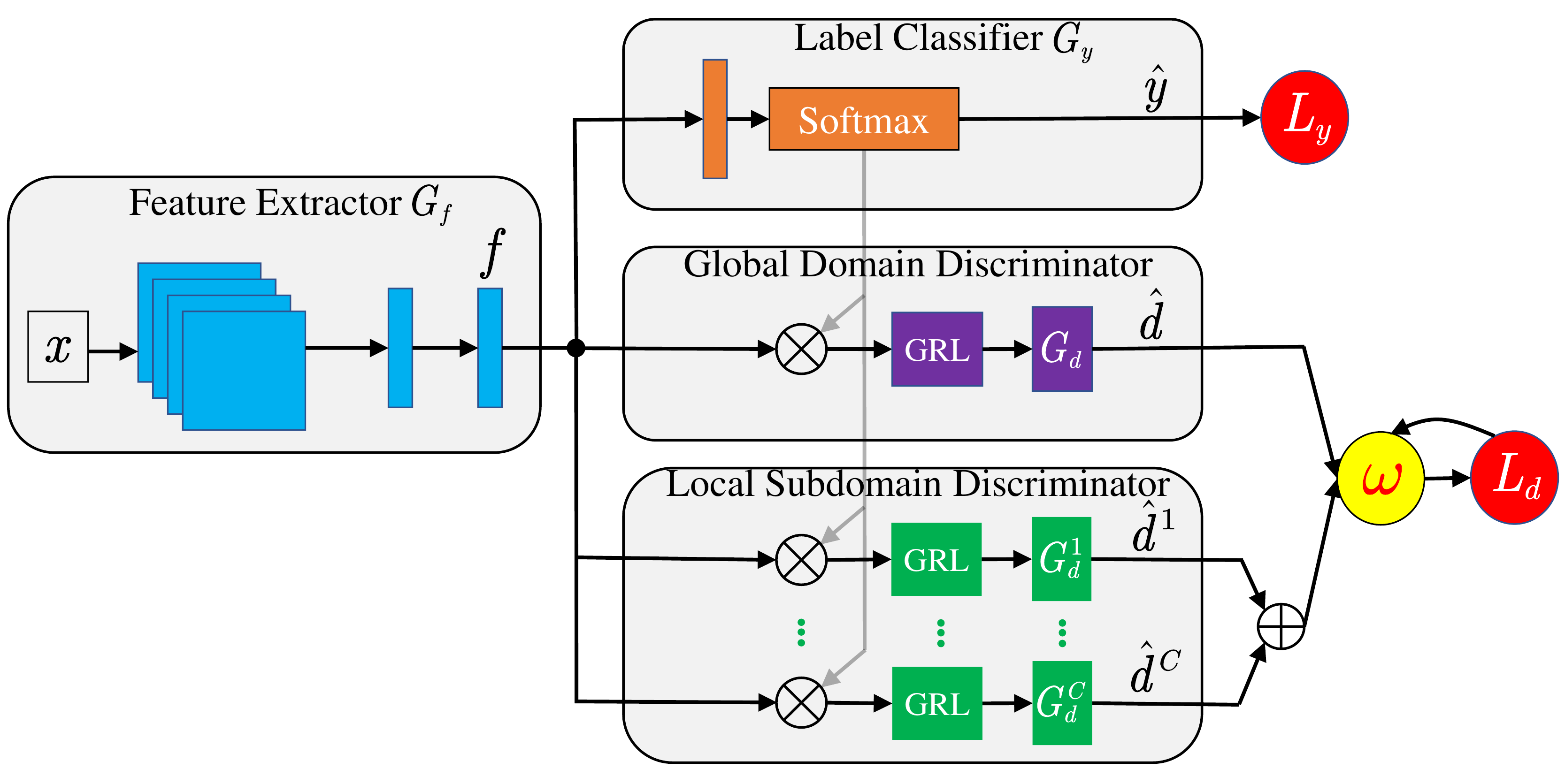}
	\vspace{-.1in}
	\caption{(Best viewed in color) The architecture of the proposed Dynamic Adversarial Adaptation Network (DAAN). DAAN consists of a deep feature extractor $G_{f}$ (blue), a label classifier $G_{y}$ (orange), a global domain discriminator ($G_{d}$, purple), $C$ local subdomain discriminators ($G^{c}_{d}$ for $c \in \{1,\cdots, C\}$, green), and a dynamic adversarial factor module ($\omega$, yellow). $\oplus$ denotes the plus operator while $\otimes$ is the product operator. $f$ is the extracted deep features, $\hat{y}$ is the predicted label, $L_{y}$ and $L_{d}$ are the classification loss and domain loss. $\hat{d}$ and $\hat{d^{c}}$ are the predicted domain label. GRL stands for Gradient Reversal Layer.}
	\label{fig-framework}
	\vspace{-.15in}
\end{figure*}

\subsection{Dynamic Adversarial Adaptation Network}
\label{sec-daan}

Most state-of-the-art domain adaptation approaches~\cite{ganin2014unsupervised,pei2018multi,kumar2018co,xie2018learning} are adversarial learning methods. Competitive results are achieved by either aligning the marginal distributions~\cite{long2015learning,ganin2016domain} (Source $\rightarrow$ Target I in Fig.~\ref{fig-sub-datamc}), or aligning the conditional distributions~\cite{pei2018multi} (Source $\rightarrow$ Target II in Fig.~\ref{fig-sub-datamc}). It has been shown that aligning these two distributions together would lead to better performance~\cite{long2016deep} since both distributions are helpful in learning domain-invariant features. However, in real applications, these two different distributions may have totally different contributions to the domain discrepancy. In real applications, it is extremely challenging to account for the relative importance of these two distributions. Therefore, we need to \textit{dynamically} and \textit{quantitatively} evaluate their importance in domain adaptation.

Recently, a Manifold Embedded Distribution Alignment~(MEDA)~\cite{wang2018visual} approach has been proposed to compute the weights of marginal and conditional distributions. MEDA learns a domain-invariant classifier in the Reproducing Kernel Hilbert Space~(RKHS) while evaluating the weights of the two distributions using the proxy $\mathcal{A}$-distance~\cite{ben2007analysis}. However, MEDA has to train $1+C$ extra linear classifiers in each iteration, which is computationally expensive and time-consuming. Furthermore, MEDA can only be applied to small-scale data since it calculates the pseudo-inverse of all the samples each time thus it cannot be deployed online. To sum up, it is extremely challenging to easily, dynamically, and quantitatively evaluate the relative importance of both distributions, while the system still remains scalable to large-scale data. 

In this paper, we make key technical improvements by proposing the \textit{Dynamic Adversarial Adaptation Network~(DAAN)} to address the above challenge. As shown in Fig.~\ref{fig-framework}, DAAN is based on the well-established generative adversarial networks~(GAN)~\cite{goodfellow2014generative} that aims at learning domain-invariant features via adversarial training. In DAAN, high-level features $f$ are extracted by a feature extractor~($G_{f}$, the blue part). Then, the adaptation of marginal and conditional distributions are achieved by the \textit{Global} domain discriminator~($G_{d}$, the purple part) and \textit{Local} domain discriminator~($G_{d}^{c}$, the green part), respectively. Most importantly, DAAN proposes a novel \textit{Dynamic Adversarial Factor}~($\omega$, the yellow part) to perform easy, dynamic, and quantitative evaluation of these two distributions. Along with the label classifier~($G_y$, the orange part), the parameters of DAAN can be trained efficiently with the Gradient Reversal Layer~(GRL)~\cite{ganin2016domain}.

In the next sections, we will first introduce the label classifier, global domain discriminator, and local domain discriminator. Then, the dynamic adversarial factor is presented in Section~\ref{sec-mu}. Finally, we show the loss function of DAAN and how to train DAAN.

\subsubsection{Label Classifier}
\label{sec-cls}
The label classifier~($G_y$, the orange part in Fig.~\ref{fig-framework})~is trained to discriminate the label of the input samples from the source domain. Thus, the supervised information on $\mathcal{D}_s$ can be utilized. Its training objective is a cross-entropy loss, which can be formulated as:
\begin{equation}
	\label{equ-cls}
   L_{y} = -\frac{1}{n_{s}}\sum_{\mathbf{x}_{i} \in \mathcal{D}_{s}}\sum^{C}_{c=1}P_{\mathbf{x}_{i} \rightarrow c} \log G_{y}(G_{f}(\mathbf{x}_{i})),
\end{equation}
where $C$ is the number of classes, $P_{\mathbf{x}_{i} \rightarrow c}$ is the probability of $\mathbf{x}_{i}$ belonging to class $c$, $G_{y}$ is the label classifier and $G_{f}$ is the feature extractor.

\subsubsection{Global Domain Discriminator}
\label{sec-gdd}
The global domain discriminator~($G_d$, the purple part in Fig.~\ref{fig-framework}) is designed to align the marginal~(global) distributions between the source and target domains. The general idea of global domain discriminator follows the Domain-adversarial Neural Network~(DANN)~\cite{ganin2016domain}, which has been described in the previous section. In DAAN, we calculate the loss of the global domain discriminator as:
\begin{equation}
	\label{equ-lg}
	L_{g} = \frac{1}{n_s+n_t}\sum_{\mathbf{x}_{i} \in \mathcal{D}_{s}\cup\mathcal{D}_{t}}L_{d}(G_{d}(G_{f}(\mathbf{x}_{i})),d_{i}),
\end{equation}
where $L_{d}$ is the domain discriminator loss (cross-entropy), $G_{f}$ is the feature extractor, and $d_{i}$ is the domain label of the input sample $\mathbf{x}_{i}$.

\subsubsection{Local Domain Discriminator}
\label{sec-lsd}
The local domain discriminator~($G_{d}^{c}$, the green part in Fig.~\ref{fig-framework}) is designed to align the conditional~(local) distributions between the source and target domains. Compared to the global domain discriminator, local domain discriminator is able to align the multi-mode structure in two distributions, thus it can perform more fine-grained domain adaptation.

To be concrete, the domain discriminator $G_{d}$ can be split into $C$ \textit{class-wise} domain discriminators $G^{c}_{d}$, each is responsible for matching the source and target domain data associated with class $c$. 
The output of the label predictor $G_{y}(\mathbf{x}_{i})$ to each data point $\mathbf{x}_{i}$ can be used to indicate how much each data points $\mathbf{x}_{i}$ should be attended to the $C$ domain discriminators $G^{c}_{d}, c = 1,...,C$. The loss function of the local domain discriminator can be calculated as:
\begin{equation}
	\label{equ-ll}
	L_{l} = \frac{1}{n_s+n_t}\sum^{C}_{c=1}\sum_{\mathbf{x}_{i} \in \mathcal{D}_{s}\cup\mathcal{D}_{t}}L^{c}_{d}(G^{c}_{d}(\hat{y}^{c}_{i}G_{f}(\mathbf{x}_{i})),d_{i}),
\end{equation}
where $G^{c}_{d}$ and $L^{c}_{d}$ are the domain discriminator and its cross-entropy loss associated with class $c$, respectively. $\hat{y}^{c}_{i}$ is the predicted probability distribution over the class $c$ of the input sample $\mathbf{x}_{i}$, and $d_{i}$ is the domain label of the input sample $\mathbf{x}_{i}$.

\subsection{Dynamic Adversarial Factor $\omega$}
\label{sec-mu}
In this section, we introduce how to dynamically evaluate the global and local distributions. It is extremely challenging to design such a dynamic scheme for adversarial learning. Intuitively, there are two natural ideas to acquire $\omega$: $Random~guessing$ and $Average~search$. Random guessing randomly picks a value of $\omega$ in $[0, 1]$, then performs DAAN using the corresponding value to get the result. This process can be repeated $t$ times and the final result can be obtained by averaging all the results. Average search picks the value of $\omega=0,0.1,\cdots,1.0$ to perform DAAN $11$ times and uses the average results as the final result. However, both of these two ideas are computationally expensive for adversarial domain adaptation.

In this paper, we propose the \textit{dynamic adversarial factor}~($\omega$, the yellow part in Fig.~\ref{fig-framework}) to easily, dynamically, and quantitatively evaluate the relative importance of the marginal and conditional distributions. Compared to MEDA~\cite{wang2018visual} that needs to build $1+C$ binary classifiers for the calculation of its adaptive factor, DAAN is able to update the value of the dynamic adversarial factor within the network. Firstly, instead of using the shallow features, we use the deep adversarial representations to learn and update $\omega$, which makes DAAN more robust and accurate. Secondly, DAAN directly uses the loss of the domain discriminators to automatically fine-tune the dynamic adversarial factor, which is easier and more efficient. 

To be more specific, the global domain distributions and the local domain distributions can be seen as the marginal and conditional distributions, respectively. Therefore, in DAAN, we denote the global $\mathcal{A}$-distance of the global domain discriminator as:
\begin{equation}
	d_{\mathcal{A},g}(\mathcal{D}_{s},\mathcal{D}_{t}) = 2(1-2(L_{g})).
\end{equation}

And we calculate the local $\mathcal{A}$-distance as:
\begin{equation}
	d_{\mathcal{A},l}(\mathcal{D}^{c}_{s},\mathcal{D}^{c}_{t}) = 2(1-2(L^{c}_{l})),
\end{equation}
where $D^{c}_{s}$ and $D^{c}_{t}$ denote samples from class $c$ and $L^{c}_{l}$ is the local subdomain discriminator loss over class $c$. 
Eventually, the dynamic adversarial factor $\omega$ can be estimated as:
\begin{equation}
	\label{equ-mu}
	\hat{\omega} = \frac{d_{\mathcal{A},g}(\mathcal{D}_{s},\mathcal{D}_{t})}{d_{\mathcal{A},g}(\mathcal{D}_{s},\mathcal{D}_{t}) + \frac{1}{C}\sum^{C}_{c=1}d_{\mathcal{A},l}(\mathcal{D}^{c}_{s},\mathcal{D}^{c}_{t})}.
\end{equation}

Note that there is no need to explicitly build extra classifiers in order to compute the local distances such as MEDA~\cite{wang2018visual}. In DAAN, they can be easily implemented by taking advantages of the global and local domain discriminators. More specifically, $\omega$ is initialized as $1$ in the first epoch. After each epoch, the pseudo labels of the target domain can be obtained. Then, the local distance for class $c$ can be easily computed as:
\begin{equation}
    L^{c}_{l} = CrossEntropy(\hat{\bm{d}^c},\bm{d}^c),
\end{equation}
where $\hat{\bm{d}^c} = [\hat{\bm{d}^c_s};\hat{\bm{d}^c_t}]$ is the concatenation of the predictions output by the $c$-th domain discriminator $d_c$, and $\bm{d}^c = [\mathbf{0};\mathbf{1}]$ with $\mathbf{0} \in \mathbb{R}^{|\hat{\bm{d}^c_s}| \times 1}$ and $\mathbf{1} \in \mathbb{R}^{|\hat{\bm{d}^c_t}| \times 1}$ is the concatenation of the true domain labels (suppose the source domain has label $0$ and target domain has label $1$). Similarly, the global distances can be obtained. The calculation of the dynamic adversarial factor can be performed after each epoch of iteration. Eventually, DAAN will learn a rather robust dynamic adversarial factor as the training converges.

\subsection{Learning Procedure}

DAAN mainly consists of three components: Label Classifier~(Eq.~(\ref{equ-cls})), Global Domain Discriminator~(Eq.~(\ref{equ-lg})), and Local Subdomain Discriminator~(Eq.~(\ref{equ-ll})). Integrating all components, the learning objective of DAAN can finally be formulated as:
\begin{equation}
	\label{equ-final}
	L(\theta_{f},\theta_{y},\theta_{d},\theta^{c}_{d}|^{C}_{c=1}) = L_{y} - \lambda((1-\omega)L_{g} + \omega L_{l}),
\end{equation}
where $\lambda$ is a trade-off parameter.

It is worth noting that although DAAN involves two hyperparameters ($\lambda$ and $\omega$), the value of $\omega$ can be \textit{self-calculated} by the network. Therefore, DAAN remains the same sample and efficient as other popular adversarial methods~\cite{ganin2016domain,long2016deep}.

When $\omega \rightarrow 0$, it means that the global distribution alignment is more important~(Target I in Fig.~\ref{fig-sub-datamc}), and DAAN will degenerate to DANN~\cite{ganin2016domain}. 
When $\omega \rightarrow 1$, it means that global distributions between two domains are relatively small, so the local subdomain distributions of each class is dominant~(Target II in Fig.~\ref{fig-sub-datamc}). In this case, DAAN will degenerate to MADA~\cite{pei2018multi}. Note that in real applications, the marginal and conditional distributions are not determined. Therefore, by learning the dynamic adversarial factor $\omega$, DAAN can be applied to diverse domain adaptation scenarios.

Denoting $\bm{\Theta}=\{\theta_{f},\theta_{y},\theta_{d},\theta^{c}_{d}|^{C}_{c=1}\}$ as all the parameters to be learned, the gradient of Eq.~(\ref{equ-final}) can be computed as:
\begin{equation}
    \label{eq-gridient}
    \Delta_{\bm{\Theta}} = \frac{\Delta L_y}{\Delta \bm{\Theta}} - \lambda \frac{\Delta ((1-\omega)L_{g} + \omega L_{l})}{\Delta \bm{\Theta}}
\end{equation}

DAAN can be trained efficiently by the Stochastic Gradient Descent~(SGD) algorithm. There are two alternatives for the optimization of DAAN. We can either optimize Eq.~(\ref{eq-gridient}) directly according to~\cite{ganin2016domain}, or we can optimize the two objectives in Eq.~(\ref{eq-argmin}) iteratively.

\subsection{Discussions}

In theory, the risk of DAAN can be bounded by the following theorem since it is designed by directly minimizing the target risk according to ~\cite{ben2007analysis}:

\textbf{Theorem 1} \textit{Let $h \in \mathcal{H}$ be a hypothesis, $\epsilon_s(h)$ and $\epsilon_t(h)$ be the expected risks on the source and target domain, respectively, then}
\begin{equation}
    \label{eq-theorem}
    \epsilon_{t}(h) \leqslant \epsilon_{s}(h)+d_{\mathcal{H}}(p, q)+C_{0},
\end{equation}
where $C_0$ is a constant for the complexity of hypothesis and plus the risk of an ideal hypothesis for both domains. More importantly, according to \cite{ben2007analysis}, $d_{\mathcal{H}}(p, q)$ is $\mathcal{H}$-divergence between domains, which can be approximately measured by the $\mathcal{A}$-distances in Eq.~(\ref{equ-lg}) and Eq.~(\ref{equ-ll}). In fact, we can regard the dynamic distribution adaptation of DAAN as the dynamic version of $\mathcal{H}$-divergence, which could learn both global and local divergences between domains. Therefore, the risk of DAAN can be theoretically bounded.

DAAN can also be explained using the attention mechanism~\cite{zagoruyko2016paying}. Attention plays a critical role in human visual experience. In a computer vision task, attention tries to learn the important factors of the images. In a machine translation task, attentions helps to learn the important hidden states of the encoders. In transfer learning, we can regard that DAAN is learning the dynamic importance of the marginal and conditional distributions using neural networks. Therefore, it perceives the accurate information about distributions using the dynamic adversarial factor.

DAAN is significantly different from existing adversarial adaptation methods. Specifically, compared with global domain adaptation methods~\cite{ganin2014unsupervised,ganin2016domain} and the local subdomain adaptation methods~\cite{pei2018multi}, DAAN is able to perform dynamic adversarial distribution alignment by quantitatively calculating the relative importance of global and local distributions with a novel dynamic adversarial factor $\omega$. Compared with MEDA~\cite{wang2018visual}, DAAN uses deep adversarial representations to fine-tune $\omega$ without training extra classifiers, which makes our estimation of $\omega$ significantly more accurate, easy, and efficient.

\begin{figure}[t!]
    \centering
    \includegraphics[scale=.25]{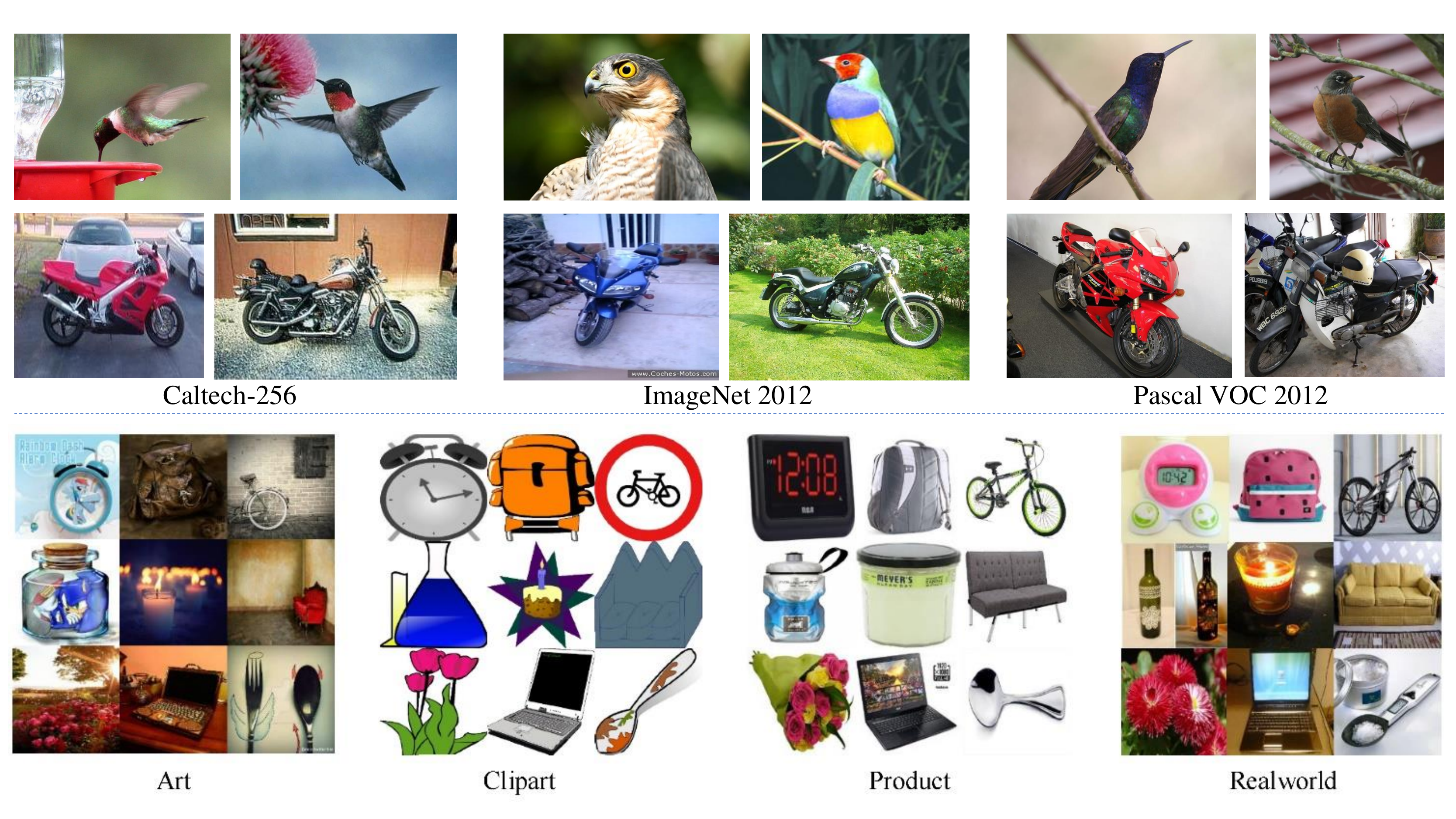}
    \caption{Datasets. Up: ImageCLEF-DA; Down: Office-Home}
    \label{fig-dataset}
\end{figure}

\begin{table*}[t!]
	\centering
	\caption{\upshape Accuracy(\%) on Office-Home for unsuperevised domain adaptation.}
	\label{tb-officehome}
	\resizebox{18cm}{!}{%
	\begin{tabular}{cccccccccclccc}
		\toprule
		Method & A$\rightarrow$C  & A$\rightarrow$P  & A$\rightarrow$R  & C$\rightarrow$A  & C$\rightarrow$P  & C$\rightarrow$R  & P$\rightarrow$A  & P$\rightarrow$C  & P$\rightarrow$R  & R$\rightarrow$A  & R$\rightarrow$C  & R$\rightarrow$P  & AVG  \\ \hline
		ResNet~\cite{he2016deep} & 34.9 & 50.0 & 58.0 & 37.4 & 41.9 & 46.2 & 38.5 & 31.2 & 60.4 & 53.9 & 41.2 & 59.9 & 46.1 \\
		DAN~\cite{long2015learning}    & 43.6 & 57.0 & 67.9 & 45.8 & 56.5 & 60.4 & 44.0 & 43.6 & 67.7 & 63.1 & 51.5 & 74.3 & 56.3 \\
		DANN~\cite{ganin2014unsupervised}   & 45.6 & 59.3 & 70.1 & 47.0 & 58.5 & 60.9 & 46.1 & 43.7 & 68.5 & 63.2 & 51.8 & 76.8 & 57.6 \\
		JAN~\cite{long2016deep}    & 45.9 & 61.2 & 68.9 & 50.4 & 59.7 & 61.0 & 45.8 & 43.4 & 70.3 & 63.9 & 52.4 & 76.8 & 58.3 \\
		MEDA~\cite{wang2018visual}   & 46.6 & \textbf{68.9} & 68.8 & 49.0 & \textbf{66.4} & \textbf{66.1} & 51.8 & 45.0 & 72.9 & 61.2 & 50.3 & 76.0 & 60.2 \\ \hline
		DAAN   & \textbf{50.5} & 65.0 & \textbf{73.7} & \textbf{53.7} & 62.7 & 64.6 & \textbf{53.5} & \textbf{45.2} & \textbf{74.0} & \textbf{66.3} & \textbf{54.0} & \textbf{78.8} & \textbf{61.8} \\ \bottomrule
	\end{tabular}%
	}
\end{table*}

\section{Experiments}
\label{sec-exp}
In this section, we evaluate the proposed DAAN against several state-of-the-art transfer learning methods on unsupervised domain adaptation problems. DAAN is validated on two popular datasets: ImageCLEF-DA~\cite{long2016deep} and Office-Home~\cite{venkateswara2017deep}. The code of DAAN is released at \url{http://transferlearning.xyz}.

\subsection{Datasets}

Examples of the two datasets are shown in Figure~\ref{fig-dataset}.

\textbf{ImageCLEF-DA} is a benchmark dataset for ImageCLEF 2014 domain adaptation challenge, and it is collected by 
selecting the 12 common categories shared by the following public datasets and each of them is considered as a domain: $Caltech-256$ (\textbf{C}), $ImageNet~ILSVRC~2012$ (\textbf{I}), $Pascal~VOC~2012$ (\textbf{P}). There are 50 images in each category and 600 images in each domain. We use all domain combinations and build 6 transfer tasks: \textbf{I}$\to$\textbf{P}, \textbf{P}$\to$\textbf{I}, \textbf{I}$\to$\textbf{C}, \textbf{C}$\to$\textbf{I}, \textbf{P}$\to$\textbf{C} and \textbf{C}$\to$\textbf{P}.

\textbf{Office-Home} is a new dataset which consists 15,588 images, which is much larger than ImageCLEF-DA. It consists of images from 4 different domains: $Artistic~images$ (\textbf{A}), $Clip~Art$ (\textbf{C}), $Product~images$ (\textbf{P}) and $Real-World~images$ (\textbf{R}). For each domain, the dataset contains images of 65 object categories collected in office and home settings. Similarly, we use all domain combinations and construct 12 transfer tasks.

\subsection{Baselines}
We compare our proposed Dynamic Adversarial Adaptation Network~(DAAN) with several state-of-the-art deep unsupervised domain adaptation methods: 
\begin{itemize}
    \item Deep residual learning~\cite{he2016deep}
    \item Deep Domain Confusion (\textbf{DDC})~\cite{tzeng2014deep}
    \item Deep Adaptation Network (\textbf{DAN})~\cite{long2015learning}
    \item Residual Transfer Network (\textbf{RTN})~\cite{long2016unsupervised}
    \item Domain Adversarial Neural Networks (\textbf{DANN})~\cite{ganin2014unsupervised}
    \item Deep CORAL (\textbf{D-CORAL})~\cite{sun2016deep}
    \item Joint Adaptation Networks (\textbf{JAN})~\cite{long2016deep}
    \item Multi-Adversarial Domain Adaptation (\textbf{MADA})~\cite{pei2018multi}
    \item Collaborative and Adversarial Network (\textbf{CAN})~\cite{zhang2018collaborative}
    \item Manifold Embedded Distribution Alignment (\textbf{MEDA})~\cite{wang2018visual}
\end{itemize}

\subsection{Implementation Details}
We implement all deep methods based on the PyTorch~\cite{paszke2017automatic} framework, and fine-tune from ResNet-50~\cite{he2016deep} models pre-trained on the ImageNet dataset~\cite{russakovsky2015imagenet}. We obtain the results of MEDA by running it on the features pre-trained by ResNet. For all the unsupervised domain adaptation tasks, we fine-tune all convolutional and pooling layers and train the classifier layer via backpropagation. Since the classifier is trained from scratch, we set its learning rate to be 10 times that of the other layers. The mini-batch Stochastic Gradient Descent~(SGD) with momentum of 0.9 is taken as optimization scheme, and the learning rate changing strategy follows existing work~\cite{ganin2014unsupervised}: the learning rate is not selected by a grid search due to high computational cost, it is adjusted during SGD using these formulas~\cite{ganin2016domain}: $\eta_{k} = \frac{\eta_{0}}{(1+\alpha k)^{\beta}}$, where $k$ is the training progress linearly changing from 0 to 1, $\eta_{0}=0.01$, $\alpha=10$ and $\beta=0.75$. We fix $\lambda=1,batchsize=32$ in DAAN all the time. Other Hyperparameters are tuned via transfer cross validation~\cite{zhong2010cross}. Following \cite{ganin2016domain,long2016unsupervised}, classification accuracy is used as the evaluation metric. The labels for the target domain are only used for evaluation. The results are obtained by running the method $10$ times to get the average accuracy.

\begin{table}[t!]
	\centering 
	\caption{\upshape Accuracy(\%) on ImageCLEF-DA for unsuperevised domain adaptation.}
	\label{tb-clef}
	\resizebox{.5\textwidth}{!}{%
	\begin{tabular}{cccccccc}
		\toprule
		Method  & I$\rightarrow$P            & P$\rightarrow$I            & I$\rightarrow$C            & C$\rightarrow$I            & C$\rightarrow$P            & P$\rightarrow$C            & AVG           \\ \hline
		ResNet~\cite{he2016deep}  & 74.8          & 83.9          & 91.5          & 78.0          & 65.5          & 91.2          & 80.7          \\
		DDC~\cite{tzeng2014deep}     & 74.6          & 85.7          & 91.1          & 82.3          & 68.3          & 88.8          & 81.8          \\
		DAN~\cite{long2015learning}     & 75.0          & 86.2          & 93.3          & 84.1          & 69.8          & 91.3          & 83.3          \\
		RTN~\cite{long2016unsupervised}     & 75.6          & 86.8          & 95.3          & 86.9          & 72.7          & 92.2          & 84.9          \\
		DANN~\cite{ganin2014unsupervised}    & 75.0          & 86.0          & \textbf{96.2} & 87.0          & 74.3          & 91.5          & 85.0          \\
		D-CORAL~\cite{sun2016deep} & 76.9          & 88.5          & 93.6          & 86.8          & 74.0          & 91.6          & 85.2          \\
		JAN~\cite{long2016deep}     & 76.8          & 88.0          & 94.7          & \textbf{89.5} & 74.2          & 91.7          & 85.8          \\
		MADA~\cite{pei2018multi}    & 75.0          & 87.9          & 96.0          & 88.8          & 75.2          & 92.2          & 85.8          \\
		CAN~\cite{zhang2018collaborative}     & 78.2          & 87.5          & 94.2          & \textbf{89.5} & \textbf{75.8} & 89.2          & 85.7          \\ 
		MEDA~\cite{wang2018visual}    & 78.1    & 90.4    & 93.1    & 86.4    & 73.2    & 91.7    & 85.5    \\ \hline
		DAAN    & \textbf{78.5} & \textbf{91.3} & 94.4          & 88.4          & 74.0          & \textbf{94.3} & \textbf{86.8} \\ \bottomrule
	\end{tabular}%
	}
\end{table}

\begin{table*}[t!]
	\centering
	\caption{\upshape Error comparison between our evaluation of $\omega$ and Average search and MEDA (the results of grid search are 0)}

	\label{tb-compareerr}
	\begin{tabular}{ccccccccc|c}
        \toprule
        Task & I $\rightarrow$ P & P$\rightarrow$I & I$\rightarrow$C & C$\rightarrow$I & A$\rightarrow$R & R$\rightarrow$A & A$\rightarrow$C & C$\rightarrow$A & AVG \\ \hline
        \begin{tabular}[c]{@{}c@{}}Avg search\\ \end{tabular} & 1.50 & 2.30 & 1.70 & 2.51 & 1.57 & 1.20 & 1.32 & 2.00 & 1.76 \\ 
        \begin{tabular}[c]{@{}c@{}}MEDA\\ \end{tabular} & 0.48 & 0.67 & 0.34 & 0.50 & 1.49 & 0.89 & 0.57 & 1.32 & 0.78 \\ 
        \textbf{DAAN} & \textbf{0.15} & \textbf{0.24} & \textbf{0.1} & \textbf{0.32} & \textbf{0.20} & \textbf{0.29} & \textbf{0.42} & \textbf{0.36} & \textbf{0.26} \\ \bottomrule
        \end{tabular}%
	\vspace{-.1in}
\end{table*}

\subsection{Results}
The classification accuracy on the ImageCLEF-DA dataset based on ResNet is shown in Table~\ref{tb-clef}. DAAN outperforms all comparison methods on most transfer tasks. It is noteworthy that DAAN outperforms DANN and MADA. Table~\ref{tb-officehome} shows the results of DAAN and several baselines on the more challenging Office-Home dataset. DAAN also outperforms all comparison methods on most tasks. 

Combining these results, more insightful conclusions can be made. \textbf{(1)}~The adversarial based methods~(DANN~\cite{ganin2014unsupervised}, MADA~\cite{pei2018multi}, and our DAAN) usually perform better than the non-adversarial based methods~(DDC~\cite{tzeng2014deep}, DAN~\cite{long2015learning}, RTN~\cite{long2016unsupervised}), which indicates that domain-adversarial learning is important for domain adaptation. 
\textbf{(2)}~Compared with the recent domain adaptation methods MEDA and CAN, DAAN achieves better performance which proves our method is more effective. 
\textbf{(3)}~In contrast to other latest adversarial methods, especially DANN~\cite{ganin2014unsupervised} and MADA~\cite{pei2018multi}, our DAAN shows better performance. During training, we also noticed that our DAAN is able to converge quickly (within < 30 epoches) compared to other methods. This implies its fast training performance with steady results (which will be shown in later experiments). This indicates that DAAN is capable of performing dynamic adversarial distribution alignment by quantitatively calculating the relative importance of global and local distributions. 

\subsection{Analysis of the Importance of the Dynamic Adversarial Factor $\omega$}

In this section, we evaluate the importance of dynamic adversarial factor $\omega$ in DAAN. To this end, there are two questions to be answered: 1)~Is it necessary to consider the different effects of marginal and conditional distributions in adversarial domain adaptation? And 2)~Is our evaluation method of $\omega$ effective?

\begin{figure}[t!]
	\centering\includegraphics[scale=0.46]{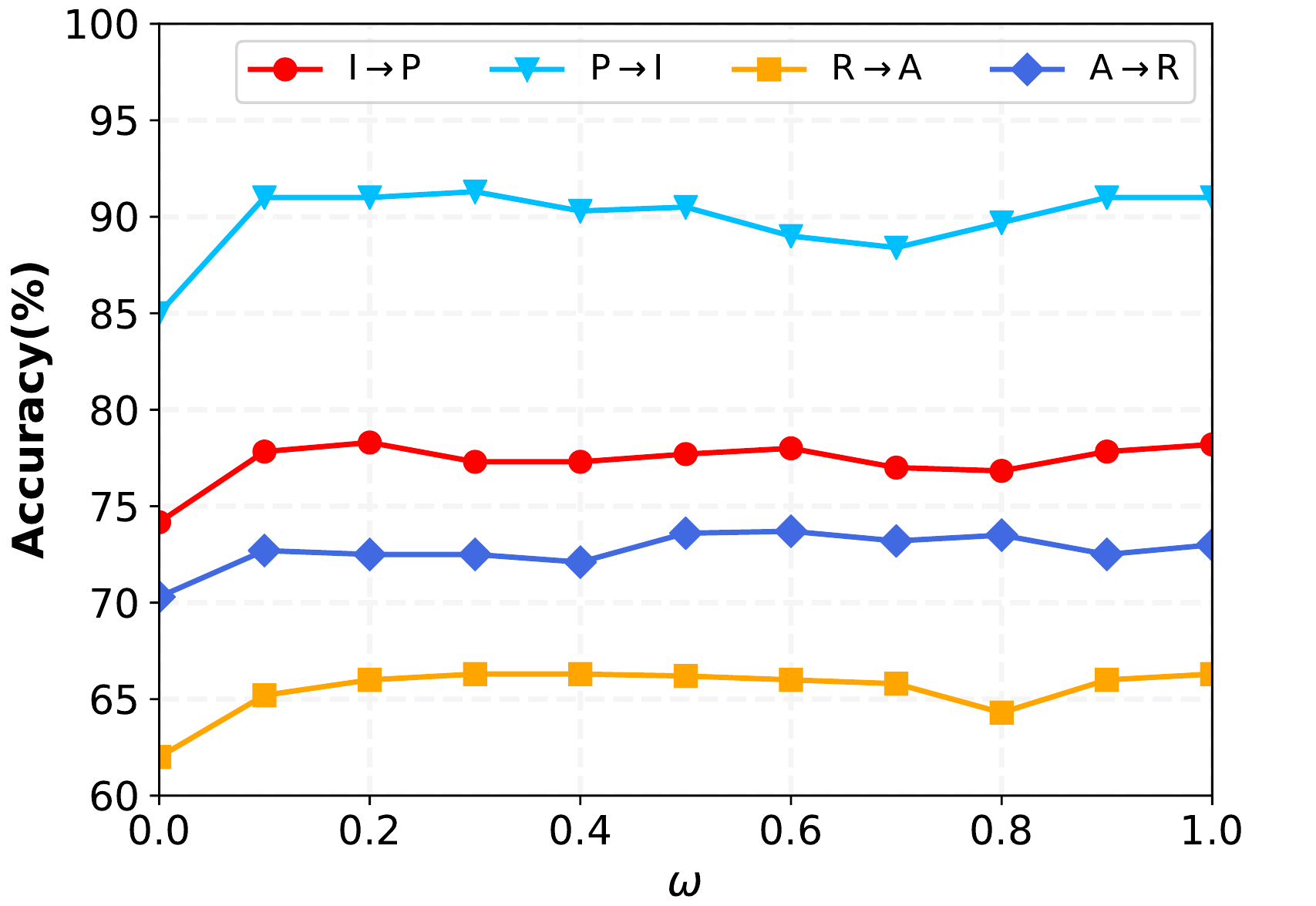}
	\vspace{-.0in}
	\caption{Performance of several tasks when searching $\omega \in [0,1]$.}
	\label{fig-differentmu}
    \vspace{-.1in}
\end{figure}

To answer the first question, we randomly pick two tasks from Office-Home and ImageCLEF-DA and draw the results of DAAN under different $\omega$ in Fig.~\ref{fig-differentmu}. It can be seen that the classification accuracy varies with different values of $\omega$, which indicates the \textit{necessity} to consider the different effects of the marginal~(global) distributions and conditional~(local) distributions not only in shallow domain adaptation~(which can be verified in BDA~\cite{wang2017balanced} and MEDA~\cite{wang2018visual}), but also in adversarial transfer learning. Moreover, We find that the value of optimal $\omega$ varies on different tasks and even for the same task, $\omega$ may have several optimal values. This may be because of different feature representations learned by calculating $\omega$. Again, it implies the importance of $\omega$ in adversarial domain adaptation problems.

\begin{figure}[t!]
	\centering\includegraphics[scale=0.48]{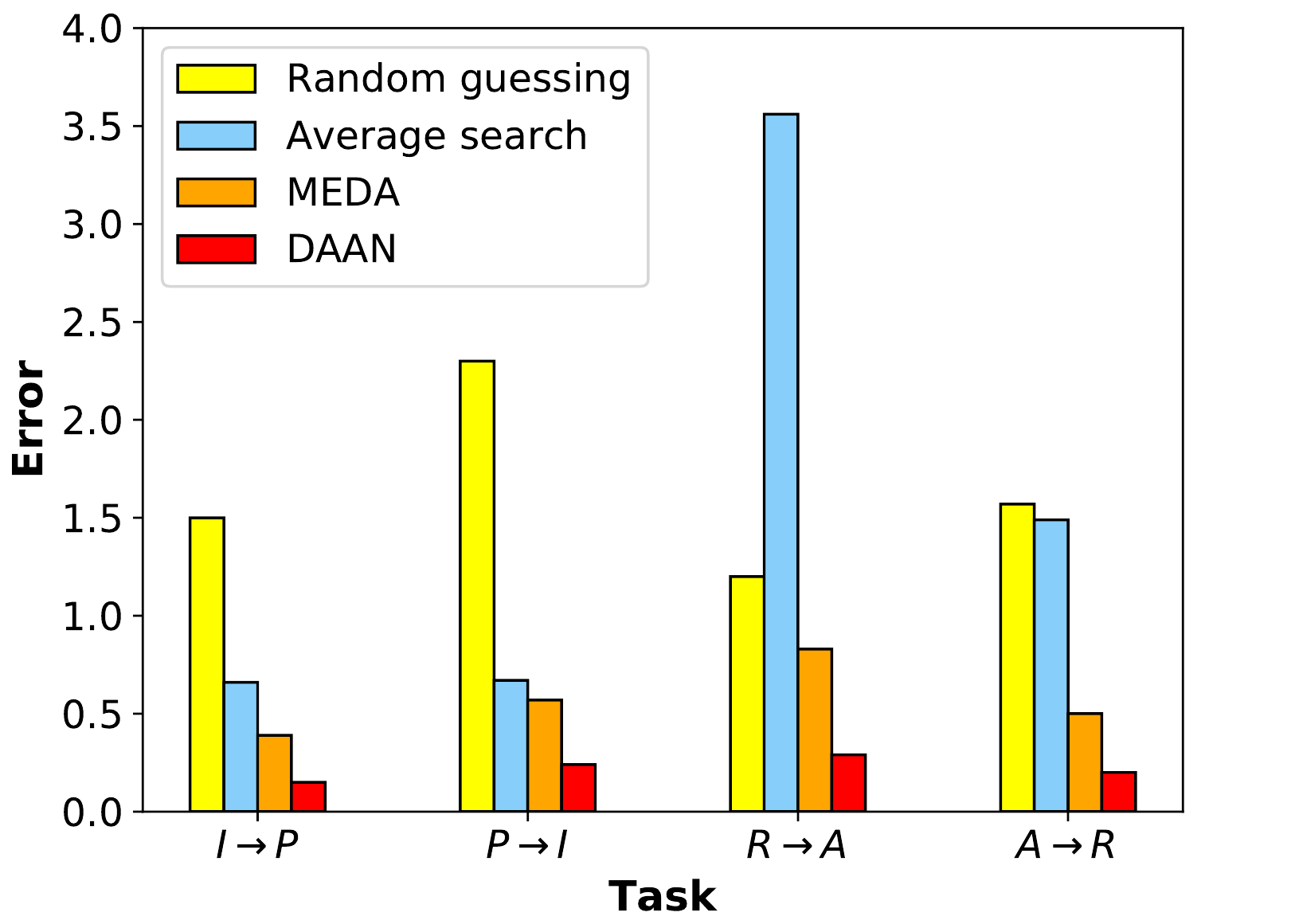}
	\caption{(Best viewed in color) The performance w.r.t different calculation method of $\omega$: Random guessing, Average search, MEDA, and our DAAN}
	\label{fig-differr}
	\vspace{-.1in}
\end{figure}

To answer the second question, we compare the accuracy of domain adaptation tasks contributed by different calculation method of $\omega$: Random guessing ($t=20$), Average search, MEDA~\cite{wang2018visual}, and our DAAN. For a fair study, the results of MEDA is obtained by replacing the dynamic adversarial factor in DAAN with the adaptive factor in MEDA. Note that there is \textit{no} ground truth for $\omega$. Instead, we run DAAN and record its accuracy by \textit{grid search} $\omega \in \{0,0.1,...,0.9,1.0\}$ to find the optimal results as the ground truth. We use the labels of the target domain only for evaluation. The results are shown in Fig.~\ref{fig-differr}. Additionally, we also list some results in Table~\ref{tb-compareerr} to show the errors of different calculation methods. Combining the results, we can conclude that our evaluation of $\omega$ significantly and consistently outperforms other comparison methods. In addition, our evaluation is more efficient than the other three methods since it only requires to run the whole network \textit{once} while other methods require to run DAAN several times to get stable results. Compared with MEDA, our evaluation is more efficient and accurate since it uses the adversarial representations and does not need to train extra linear classifiers. Furthermore, it is worth noting that our evaluation of $\omega$ is extremely \textit{close} to the grid search results (which can never be reached in real applications). Therefore, the proposed dynamic adversarial factor is necessary and our evaluation of $\omega$ is more effective and efficient.

\subsection{Effectiveness Analysis}

In this section, we analyze the effectiveness of DAAN from several aspects: ablation study, feature visualizations, and convergence analysis.

\subsubsection{Ablation Study}

We compare the performance of DAAN with DANN~($\omega = 0$), MADA~($\omega = 1$), and JAN~($\omega = 0.5$). All these methods can be seen as special cases of our DAAN. The average results on each dataset in Table~\ref{tb-ablation} indicate that it is not enough to only align the marginal or conditional distributions, or aligning them with equal weights. Therefore, the proposed DAAN is able to perform dynamic distribution alignment between domain and achieve better performance. This property of DAAN is extremely important in real applications since given an unknown target domain, we can never know the contributions of either marginal or conditional distribution in domain divergence. DAAN makes it possible to easily, dynamically, and quantitatively evaluate their relative importance in adversarial learning.

\begin{figure}[t!]
	\centering
	\subfigure[JAN: P $\rightarrow$ I]{
		\centering
		\includegraphics[scale=0.302]{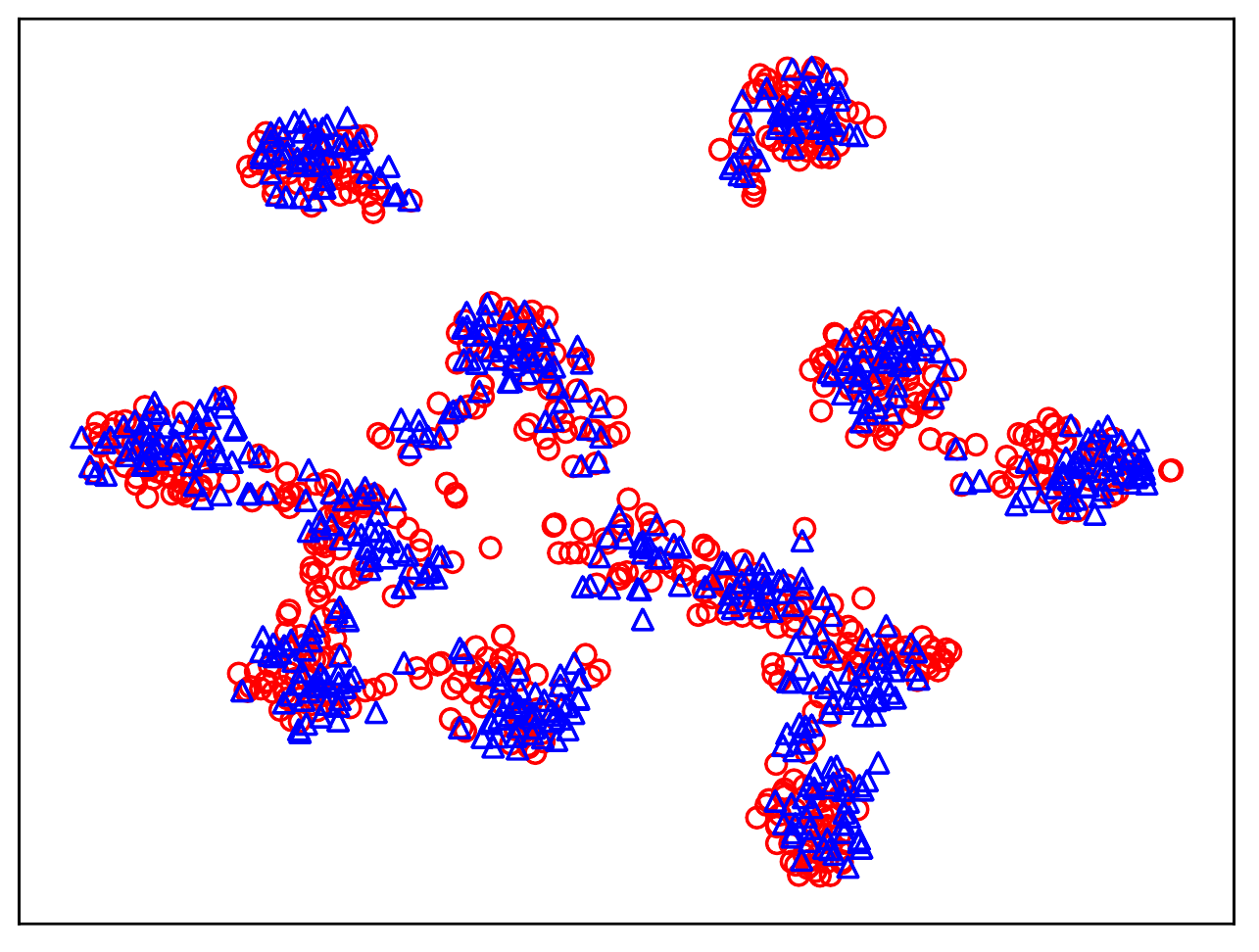}
		\label{fig-sub-dan}}
	\subfigure[DAAN: P $\rightarrow$ I]{
		\centering
		\includegraphics[scale=0.302]{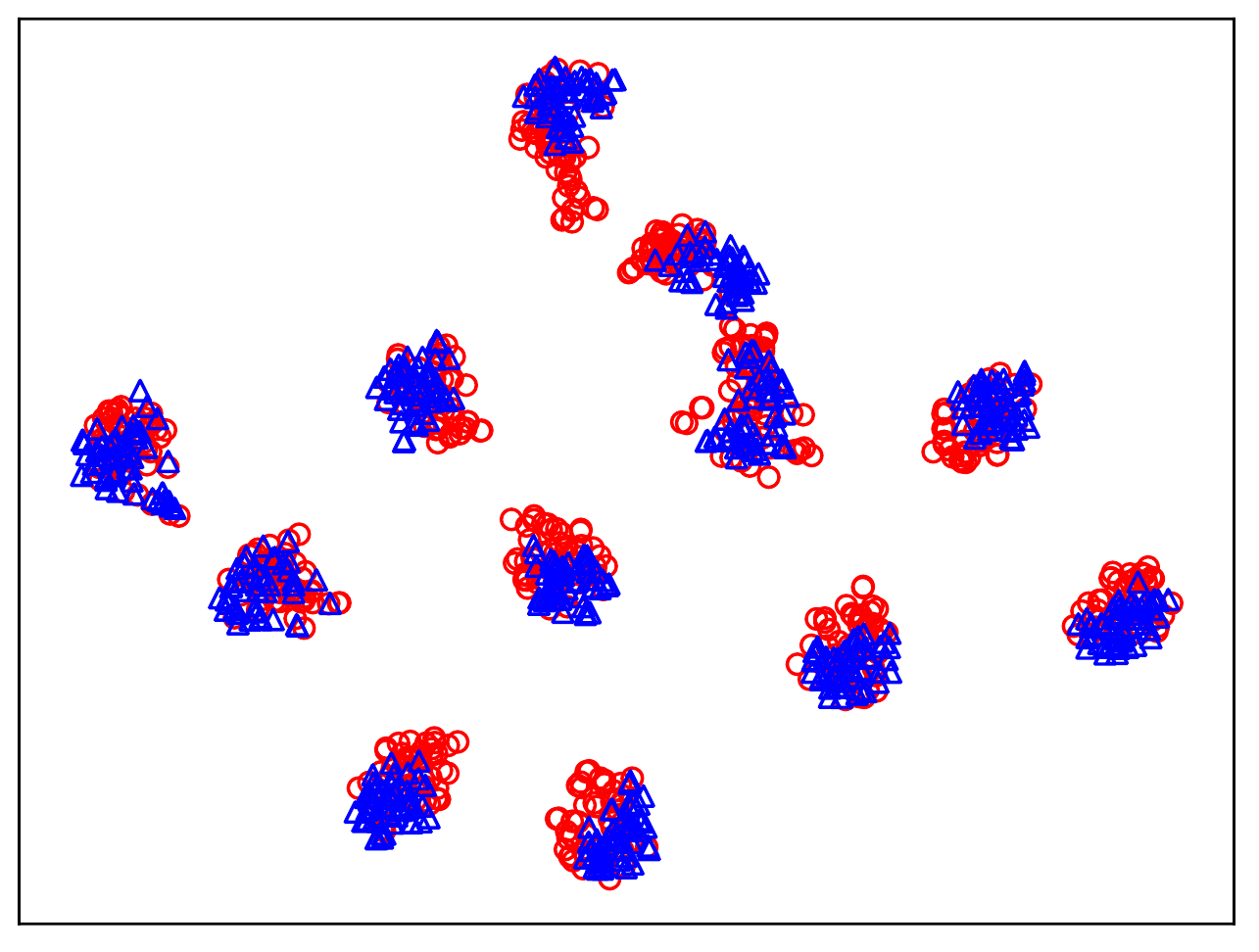}
		\label{fig-sub-daan}}
	\caption{(Best viewed in color) The t-SNE visualization of network activation. (a) and (b) are the learned representations on task P $\rightarrow$ I, respectively.}
	\label{fig-awresinfo}
	\vspace{-.0in}
\end{figure}

\begin{table}[t!]
\caption{\upshape Ablation study of DAAN}
\label{tb-ablation}
\resizebox{.48\textwidth}{!}{
\begin{tabular}{ccccc}
\toprule
Dataset & \begin{tabular}[c]{@{}c@{}}DANN\\ ($\omega=1$)\end{tabular} & \begin{tabular}[c]{@{}c@{}}MADA\\ ($\omega=0$)\end{tabular} & \begin{tabular}[c]{@{}c@{}}JAN\\ ($\omega=0.5$)\end{tabular} & DAAN \\ \hline
ImageCLEF-DA & 85.0 & 85.8 & 85.8 & \textbf{86.8} \\ 
Office-Home & 57.6 & - & 58.3 & \textbf{61.8} \\ \bottomrule
\end{tabular}}
\end{table}

\subsubsection{Feature Visualization}
To further evaluate the performance of DAAN, we visualize the network activations on task P $\rightarrow$ I~(12 classes) learned by JAN and DAAN using t-SNE embeddings~\cite{donahue2014decaf} in Fig.~\ref{fig-sub-dan}-\ref{fig-sub-daan}. Red circles are the source samples and blue triangles are the target samples. The visualization results reveal some important observations. \textbf{(1)}~As for the results of JAN, the distributions between the source and target domains are not aligned very well and different categories are not well discriminated clearly. \textbf{(2)}~In contrast, for the representations learned with our DAAN, not only the distributions between the source and target domains are aligned very well, different categories can also be discriminated more clearly. This ensures that our DAAN can achieve better performance. 
The above observations suggest that DAAN is able to learn more representative and transferable features by quantitatively calculating the relative importance of global domain distributions and local subdomain distributions. 

\begin{figure}[t!]
	\centering
	\subfigure[$\omega$]{
		\centering
		\includegraphics[scale=0.28]{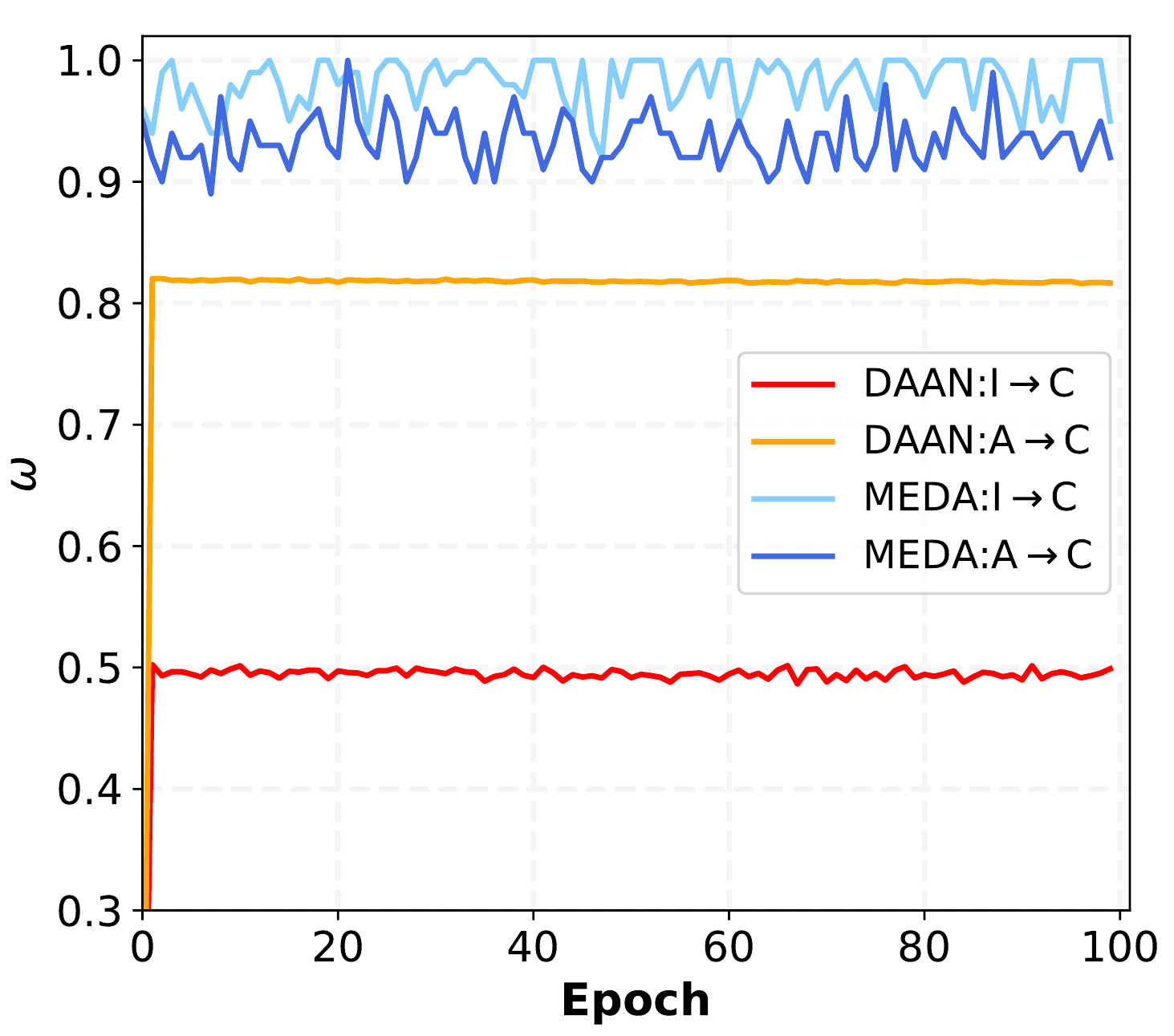}
		\label{fig-sub-mu}}
	\hspace{-.1in}
	\subfigure[Loss]{
		\centering
		\includegraphics[scale=0.28]{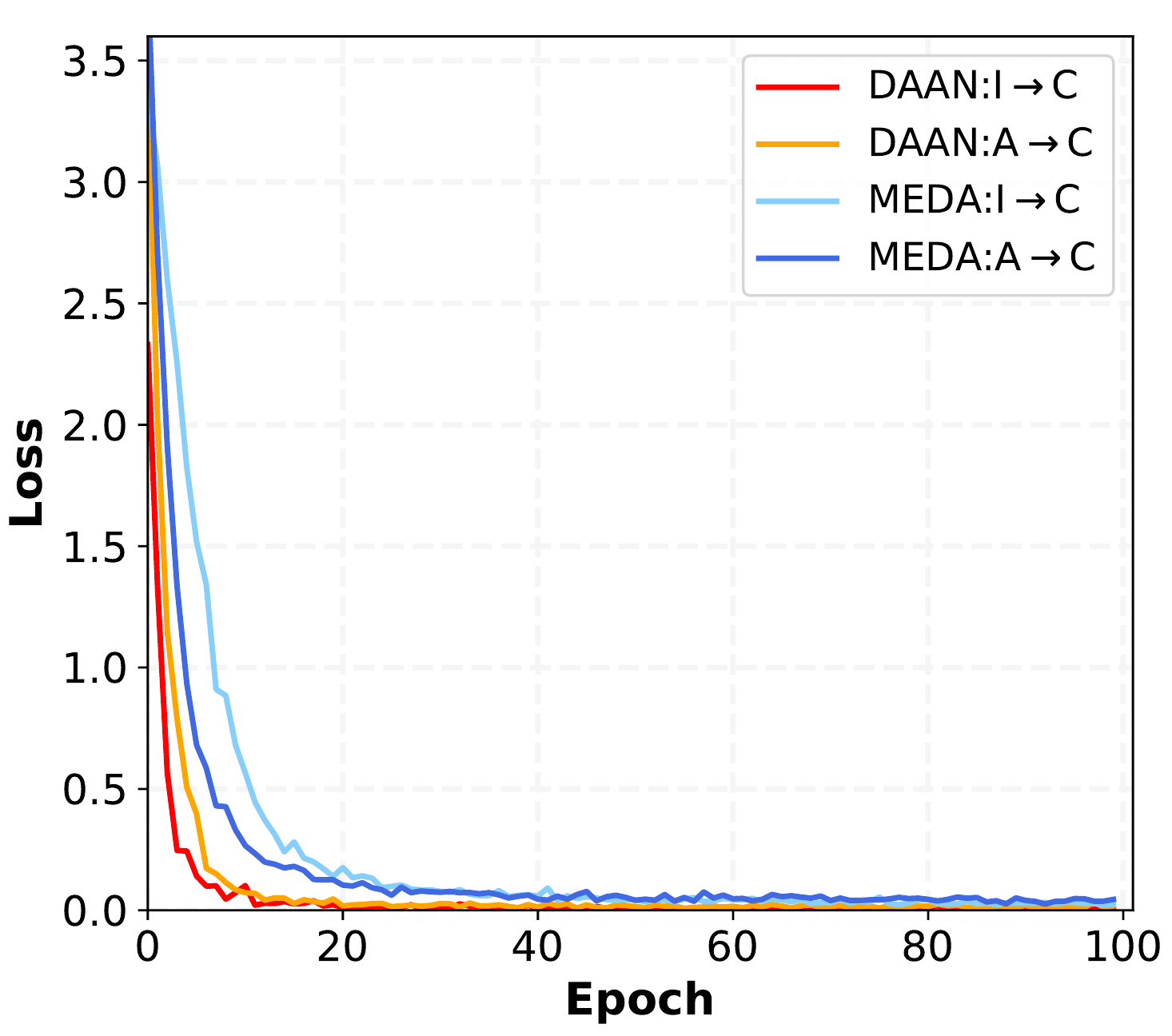}
		\label{fig-sub-loss}}
	\caption{Value change of dynamic adversarial factor $\omega$ and loss w.r.t. iterations}
	\label{fig-mu-loss}
	\vspace{-.1in}
\end{figure}

\subsubsection{Convergence Analysis}

In this section, we evaluate the convergence of DAAN. On the same DAAN architecture, we compare the change of $\omega$ between DAAN and MEDA w.r.t. iterations in Fig.~\ref{fig-sub-mu}. Additionally, their loss can be seen in Fig.~\ref{fig-sub-loss}. From these results, we can observe: \textbf{(1)}~DAAN can reach a quick and steady convergence after 20 epochs. \textbf{(2)}~The dynamic adversarial factor $\omega$ can also reach a steady value after several iterations, while the adaptive factor of MEDA takes more iterations. These results demonstrate that the proposed DAAN can not only reach competitive performances, it can also be trained easily with steady results.

\section{Conclusions and Future Work}
\label{sec-conclu}
In this paper, we propose a novel Dynamic Adversarial Adaptation Network~(DAAN) for adversarial transfer learning. DAAN is able to learn a domain-invariant network while performing dynamic adversarial distribution alignment to quantitatively evaluate the relative importance of marginal (global) domain distributions and conditional (local) subdomain distributions. 
To the best of our knowledge, DAAN is the first attempt to perform easy, dynamic, and quantitative evaluation of these two distributions in adversarial neural networks.
DAAN can be easily implemented and used in real domain adaptation tasks. Experimental results demonstrate that DAAN achieves superior performance compared to state-of-the-art deep methods.

DAAN is a general transfer learning and domain adaptation approach and it can be applied to a large amount of classification related applications such as object detection, image segmentation, and visual tracking. In the future, we plan to extend DAAN for the more challenging cross-domain data mining problems.

\section{Acknowledgment}
The first two authors contributed equally. This work is supported in part by National Key R \& D Plan of China~(No.2017YFB1002802), National Natural Science Foundation of China~(No.61572471), Beijing Municipal Science \& Technology Commission~(No.Z171100000117001) and Chinese Academy of Sciences Research Equipment Development Project under Grant~(No.YZ201527).

\bibliographystyle{IEEEtran}
\bibliography{icdm19}

\begin{thebibliography}{10}
\providecommand{\url}[1]{#1}
\csname url@samestyle\endcsname
\providecommand{\newblock}{\relax}
\providecommand{\bibinfo}[2]{#2}
\providecommand{\BIBentrySTDinterwordspacing}{\spaceskip=0pt\relax}
\providecommand{\BIBentryALTinterwordstretchfactor}{4}
\providecommand{\BIBentryALTinterwordspacing}{\spaceskip=\fontdimen2\font plus
\BIBentryALTinterwordstretchfactor\fontdimen3\font minus
  \fontdimen4\font\relax}
\providecommand{\BIBforeignlanguage}[2]{{%
\expandafter\ifx\csname l@#1\endcsname\relax
\typeout{** WARNING: IEEEtran.bst: No hyphenation pattern has been}%
\typeout{** loaded for the language `#1'. Using the pattern for}%
\typeout{** the default language instead.}%
\else
\language=\csname l@#1\endcsname
\fi
#2}}
\providecommand{\BIBdecl}{\relax}
\BIBdecl

\bibitem{he2016deep}
K.~He, X.~Zhang, S.~Ren, and J.~Sun, ``Deep residual learning for image
  recognition,'' in \emph{Proceedings of the IEEE conference on computer vision
  and pattern recognition}, 2016, pp. 770--778.

\bibitem{krizhevsky2012imagenet}
A.~Krizhevsky, I.~Sutskever, and G.~E. Hinton, ``Imagenet classification with
  deep convolutional neural networks,'' in \emph{Advances in neural information
  processing systems}, 2012, pp. 1097--1105.

\bibitem{pan2010survey}
S.~J. Pan, Q.~Yang \emph{et~al.}, ``A survey on transfer learning,'' \emph{IEEE
  Transactions on knowledge and data engineering}, vol.~22, no.~10, pp.
  1345--1359, 2010.

\bibitem{huang2007correcting}
J.~Huang, A.~Gretton, K.~Borgwardt, B.~Sch{\"o}lkopf, and A.~J. Smola,
  ``Correcting sample selection bias by unlabeled data,'' in \emph{Advances in
  neural information processing systems}, 2007, pp. 601--608.

\bibitem{chen2019cross}
Y.~Chen, J.~Wang, M.~Huang, and H.~Yu, ``Cross-position activity recognition
  with stratified transfer learning,'' \emph{Pervasive and Mobile Computing},
  vol.~57, pp. 1--13, 2019.

\bibitem{wang2019easy}
J.~Wang, Y.~Chen, H.~Yu, M.~Huang, and Q.~Yang, ``Easy transfer learning by
  exploiting intra-domain structures,'' in \emph{IEEE International Conference
  on Multimedia and Expo (ICME)}, 2019.

\bibitem{wang2018visual}
J.~Wang, W.~Feng, Y.~Chen, H.~Yu, M.~Huang, and P.~S. Yu, ``Visual domain
  adaptation with manifold embedded distribution alignment,'' in \emph{2018 ACM
  International Conference on Multimedia (ACM MM)}.\hskip 1em plus 0.5em minus
  0.4em\relax ACM, 2018, pp. 402--410.

\bibitem{wang2017balanced}
J.~Wang, Y.~Chen, S.~Hao, W.~Feng, and Z.~Shen, ``Balanced distribution
  adaptation for transfer learning,'' in \emph{2017 IEEE International
  Conference on Data Mining (ICDM)}.\hskip 1em plus 0.5em minus 0.4em\relax
  IEEE, 2017, pp. 1129--1134.

\bibitem{gong2012geodesic}
B.~Gong, Y.~Shi, F.~Sha, and K.~Grauman, ``Geodesic flow kernel for
  unsupervised domain adaptation,'' in \emph{2012 IEEE Conference on Computer
  Vision and Pattern Recognition}.\hskip 1em plus 0.5em minus 0.4em\relax IEEE,
  2012, pp. 2066--2073.

\bibitem{wang2018stratified}
J.~Wang, Y.~Chen, L.~Hu, X.~Peng, and S.~Y. Philip, ``Stratified transfer
  learning for cross-domain activity recognition,'' in \emph{2018 IEEE
  International Conference on Pervasive Computing and Communications
  (PerCom)}.\hskip 1em plus 0.5em minus 0.4em\relax IEEE, 2018, pp. 1--10.

\bibitem{pan2011domain}
S.~J. Pan, I.~W. Tsang, J.~T. Kwok, and Q.~Yang, ``Domain adaptation via
  transfer component analysis,'' \emph{IEEE Transactions on Neural Networks},
  vol.~22, no.~2, pp. 199--210, 2011.

\bibitem{donahue2014decaf}
J.~Donahue, Y.~Jia, O.~Vinyals, J.~Hoffman, N.~Zhang, E.~Tzeng, and T.~Darrell,
  ``Decaf: A deep convolutional activation feature for generic visual
  recognition,'' in \emph{International conference on machine learning}, 2014,
  pp. 647--655.

\bibitem{yosinski2014transferable}
J.~Yosinski, J.~Clune, Y.~Bengio, and H.~Lipson, ``How transferable are
  features in deep neural networks?'' in \emph{Advances in neural information
  processing systems}, 2014, pp. 3320--3328.

\bibitem{zhang2018collaborative}
W.~Zhang, W.~Ouyang, W.~Li, and D.~Xu, ``Collaborative and adversarial network
  for unsupervised domain adaptation,'' in \emph{Proceedings of the IEEE
  Conference on Computer Vision and Pattern Recognition}, 2018, pp. 3801--3809.

\bibitem{zhu2019multi}
Y.~Zhu, F.~Zhuang, J.~Wang, J.~Chen, Z.~Shi, W.~Wu, and Q.~He,
  ``Multi-representation adaptation network for cross-domain image
  classification,'' \emph{Neural Networks}, 2019.

\bibitem{yu2019accelerating}
C.~Yu, J.~Wang, Y.~Chen, and Z.~Wu, ``Accelerating deep unsupervised domain
  adaptation with transfer channel pruning,'' \emph{arXiv preprint
  arXiv:1904.02654}, 2019.

\bibitem{sun2016deep}
B.~Sun and K.~Saenko, ``Deep coral: Correlation alignment for deep domain
  adaptation,'' in \emph{European Conference on Computer Vision}.\hskip 1em
  plus 0.5em minus 0.4em\relax Springer, 2016, pp. 443--450.

\bibitem{ganin2014unsupervised}
Y.~Ganin and V.~Lempitsky, ``Unsupervised domain adaptation by
  backpropagation,'' in \emph{International Conference on Machine Learning
  (ICML)}, 2015.

\bibitem{wang2018deep}
J.~Wang, V.~W. Zheng, Y.~Chen, and M.~Huang, ``Deep transfer learning for
  cross-domain activity recognition,'' in \emph{Proceedings of the 3rd
  International Conference on Crowd Science and Engineering}.\hskip 1em plus
  0.5em minus 0.4em\relax ACM, 2018, p.~16.

\bibitem{venkateswara2017deep}
H.~Venkateswara, J.~Eusebio, S.~Chakraborty, and S.~Panchanathan, ``Deep
  hashing network for unsupervised domain adaptation,'' in \emph{Proceedings of
  the IEEE Conference on Computer Vision and Pattern Recognition}, 2017, pp.
  5018--5027.

\bibitem{long2016deep}
M.~Long, H.~Zhu, J.~Wang, and M.~I. Jordan, ``Deep transfer learning with joint
  adaptation networks,'' in \emph{ICML}, 2017.

\bibitem{goodfellow2014generative}
I.~Goodfellow, J.~Pouget-Abadie, M.~Mirza, B.~Xu, D.~Warde-Farley, S.~Ozair,
  A.~Courville, and Y.~Bengio, ``Generative adversarial nets,'' in
  \emph{Advances in neural information processing systems}, 2014, pp.
  2672--2680.

\bibitem{tzeng2017adversarial}
E.~Tzeng, J.~Hoffman, K.~Saenko, and T.~Darrell, ``Adversarial discriminative
  domain adaptation,'' in \emph{Proceedings of the IEEE Conference on Computer
  Vision and Pattern Recognition}, 2017, pp. 7167--7176.

\bibitem{motiian2017few}
S.~Motiian, Q.~Jones, S.~Iranmanesh, and G.~Doretto, ``Few-shot adversarial
  domain adaptation,'' in \emph{Advances in Neural Information Processing
  Systems}, 2017, pp. 6670--6680.

\bibitem{pei2018multi}
Z.~Pei, Z.~Cao, M.~Long, and J.~Wang, ``Multi-adversarial domain adaptation,''
  in \emph{Thirty-Second AAAI Conference on Artificial Intelligence}, 2018.

\bibitem{ganin2016domain}
Y.~Ganin, E.~Ustinova, H.~Ajakan, P.~Germain, H.~Larochelle, F.~Laviolette,
  M.~Marchand, and V.~Lempitsky, ``Domain-adversarial training of neural
  networks,'' \emph{The Journal of Machine Learning Research}, vol.~17, no.~1,
  pp. 2096--2030, 2016.

\bibitem{fernando2013unsupervised}
B.~Fernando, A.~Habrard, M.~Sebban, and T.~Tuytelaars, ``Unsupervised visual
  domain adaptation using subspace alignment,'' in \emph{Proceedings of the
  IEEE international conference on computer vision}, 2013, pp. 2960--2967.

\bibitem{sun2015subspace}
B.~Sun and K.~Saenko, ``Subspace distribution alignment for unsupervised domain
  adaptation.'' in \emph{BMVC}, 2015, pp. 24--1.

\bibitem{sun2016return}
B.~Sun, J.~Feng, and K.~Saenko, ``Return of frustratingly easy domain
  adaptation.'' in \emph{AAAI}, vol.~6, no.~7, 2016, p.~8.

\bibitem{long2013transfer}
M.~Long, J.~Wang, G.~Ding, J.~Sun, and P.~S. Yu, ``Transfer feature learning
  with joint distribution adaptation,'' in \emph{Proceedings of the IEEE
  international conference on computer vision}, 2013, pp. 2200--2207.

\bibitem{zhuang2015supervised}
F.~Zhuang, X.~Cheng, P.~Luo, S.~J. Pan, and Q.~He, ``Supervised representation
  learning: Transfer learning with deep autoencoders,'' in \emph{Twenty-Fourth
  International Joint Conference on Artificial Intelligence}, 2015.

\bibitem{long2015learning}
M.~Long, Y.~Cao, J.~Wang, and M.~I. Jordan, ``Learning transferable features
  with deep adaptation networks,'' in \emph{International Conference on Machine
  Learning (ICML)}, 2015.

\bibitem{tzeng2014deep}
E.~Tzeng, J.~Hoffman, N.~Zhang, K.~Saenko, and T.~Darrell, ``Deep domain
  confusion: Maximizing for domain invariance,'' \emph{arXiv preprint
  arXiv:1412.3474}, 2014.

\bibitem{yan2017mind}
H.~Yan, Y.~Ding, P.~Li, Q.~Wang, Y.~Xu, and W.~Zuo, ``Mind the class weight
  bias: Weighted maximum mean discrepancy for unsupervised domain adaptation,''
  in \emph{Proceedings of the IEEE Conference on Computer Vision and Pattern
  Recognition}, 2017, pp. 2272--2281.

\bibitem{zellinger2017central}
W.~Zellinger, T.~Grubinger, E.~Lughofer, T.~Natschl{\"a}ger, and
  S.~Saminger-Platz, ``Central moment discrepancy (cmd) for domain-invariant
  representation learning,'' in \emph{International Conference on Learning
  Representations (ICLR)}, 2017.

\bibitem{durugkar2016generative}
I.~Durugkar, I.~Gemp, and S.~Mahadevan, ``Generative multi-adversarial
  networks,'' \emph{arXiv preprint arXiv:1611.01673}, 2016.

\bibitem{kumar2018co}
A.~Kumar, P.~Sattigeri, K.~Wadhawan, L.~Karlinsky, R.~Feris, B.~Freeman, and
  G.~Wornell, ``Co-regularized alignment for unsupervised domain adaptation,''
  in \emph{Advances in Neural Information Processing Systems}, 2018, pp.
  9367--9378.

\bibitem{xie2018learning}
S.~Xie, Z.~Zheng, L.~Chen, and C.~Chen, ``Learning semantic representations for
  unsupervised domain adaptation,'' in \emph{International Conference on
  Machine Learning}, 2018, pp. 5419--5428.

\bibitem{ben2007analysis}
S.~Ben-David, J.~Blitzer, K.~Crammer, and F.~Pereira, ``Analysis of
  representations for domain adaptation,'' in \emph{Advances in neural
  information processing systems}, 2007, pp. 137--144.

\bibitem{zagoruyko2016paying}
S.~Zagoruyko and N.~Komodakis, ``Paying more attention to attention: Improving
  the performance of convolutional neural networks via attention transfer,'' in
  \emph{International Conference on Learning Representations (ICLR)}, 2016.

\bibitem{long2016unsupervised}
M.~Long, H.~Zhu, J.~Wang, and M.~I. Jordan, ``Unsupervised domain adaptation
  with residual transfer networks,'' in \emph{Advances in Neural Information
  Processing Systems}, 2016, pp. 136--144.

\bibitem{paszke2017automatic}
A.~Paszke, S.~Gross, S.~Chintala, G.~Chanan, E.~Yang, Z.~DeVito, Z.~Lin,
  A.~Desmaison, L.~Antiga, and A.~Lerer, ``Automatic differentiation in
  pytorch,'' 2017.

\bibitem{russakovsky2015imagenet}
O.~Russakovsky, J.~Deng, H.~Su, J.~Krause, S.~Satheesh, S.~Ma, Z.~Huang,
  A.~Karpathy, A.~Khosla, M.~Bernstein \emph{et~al.}, ``Imagenet large scale
  visual recognition challenge,'' \emph{International journal of computer
  vision}, vol. 115, no.~3, pp. 211--252, 2015.

\bibitem{zhong2010cross}
E.~Zhong, W.~Fan, Q.~Yang, O.~Verscheure, and J.~Ren, ``Cross validation
  framework to choose amongst models and datasets for transfer learning,'' in
  \emph{Joint European Conference on Machine Learning and Knowledge Discovery
  in Databases}.\hskip 1em plus 0.5em minus 0.4em\relax Springer, 2010, pp.
  547--562.

\end{thebibliography}

\end{document}